\documentclass[10pt,twocolumn,letterpaper]{article}
\usepackage{cvpr} 
\makeatletter
\@namedef
{ver@everyshi.sty}{}
\makeatother
\usepackage{graphicx}
\usepackage{amsmath}
\usepackage{amssymb}
\usepackage{booktabs}

\usepackage[pagebackref,breaklinks,colorlinks,bookmarks=false]{hyperref}

\usepackage[capitalize]{cleveref}
\crefname{section}{Sec.}{Secs.}
\Crefname{section}{Section}{Sections}
\Crefname{table}{Table}{Tables}
\crefname{table}{Tab.}{Tabs.}
\usepackage{tikz}
\usepackage{times}
\usepackage{epsfig}
\usepackage{graphicx}
\usepackage{amsmath}
\usepackage{amssymb}
\usepackage[linesnumbered,ruled,lined]{algorithm2e}
\usepackage{forest}
\usetikzlibrary{arrows.meta}
\usepackage{comment} 
\usepackage{adjustbox}
\usepackage{multirow}

\usepackage[pagebackref,breaklinks,colorlinks]{hyperref}

\def\cvprPaperID{****} 
\def\confName{ICCV}
\def\confYear{2021}
\graphicspath{{./figs/}{./figs/result/}}

\tikzset{
  every leaf node/.style={draw=red,rectangle,align=center},
  every tree node/.style={draw=black,rectangle, align=center},
  tt/.style={font=\ttfamily},
}
\forestset{tikzQtree/.style={for tree={if n children=0{
        node options=every leaf node/.try}{node options=every tree node/.try}}}}
\begin{document}

\title{Neural Networks are Decision Trees}

\author{Caglar Aytekin\\
AI Lead\\
AAC Technologies\\
{\tt\small caglaraytekin@aactechnologies.com, cagosmail@gmail.com}
}

\maketitle

\begin{abstract}
   In this manuscript, we show that any neural network with any activation function can be represented as a decision tree.
   The representation is equivalence and not an approximation, thus keeping the accuracy of the neural network exactly as is. 
   We believe that this work provides better understanding of neural networks and paves the way to tackle their black-box nature.
   We share equivalent trees of some neural networks and show that besides providing interpretability, tree representation can also achieve some computational advantages for small networks.
   The analysis holds both for fully connected and convolutional networks, which may or may not also include skip connections and/or normalizations.
   
\end{abstract}

\section{Introduction}
Despite the immense success of neural networks over the past decade, the black-box nature of their predictions prevent their wider and more reliable adoption in many industries, such as health and security.
This fact led researchers to investigate ways to explain neural network decisions. 
The efforts in explaining neural network decisions can be categorized into several approaches: saliency maps, approximation by interpretable methods and joint models.

Saliency maps are ways of highlighting areas on the input, of which a neural network make use of the most while prediction.
An earlier work \cite{simonyan2013deep} takes the gradient of the neural network output with respect to the input in order to visualize an input-specific linearization of the entire network.
Another work \cite{zeiler2014visualizing} uses a deconvnet to go back to features from decisions.
The saliency maps obtained via these methods are often noisy and prevent a clear understanding of the decisions made. 
Another track of methods \cite{zhou2016learning}, \cite{selvaraju2017grad}, \cite{chattopadhay2018grad}, \cite{draelos2020use} make use of the derivative of a neural network output with respect to an activation, usually the one right before fully connected layers.
This saliency maps obtained by this track, and some other works \cite{zhang2018top}, \cite{muhammad2020eigen}, \cite{collins2018deep} are clearer in the sense of highlighting areas related to the predicted class.
Although useful for purposes such as checking whether the support area for decisions are sound, these methods still lack a detailed logical reasoning of why such decision is made.

Conversion between neural networks and interpretable by-design models -such as decision trees- has been a topic of interest. 
In \cite{humbird2018deep}, a method was devised to initialize neural networks with decision trees. 
\cite{yang2018deep,kontschieder2015deep,sethi1990entropy} also provides neural network equivalents of decision trees. The neural networks in these works have specific architectures, thus the conversion lacks generalization to any model.
In \cite{wu2018beyond}, neural networks were trained in such a way that their decision boundaries can be approximated by trees.
This work does not provide a correspondence between neural networks and decision trees, and merely uses the latter as a regularization. 
In \cite{frosst2017distilling}, a neural network was used to train a decision tree.
Such tree distillation is an approximation of a neural network and not a direct conversion, thus performs poorly on the tasks that the neural network was trained on.

Joint neural network and decision tree models \cite{mullapudi2018hydranets},\cite{redmon2017yolo9000},\cite{murdock2016blockout},\cite{murthy2016deep},\cite{roy2016monocular},\cite{ahmed2016network},\cite{mcgill2017deciding}, \cite{veit2018convolutional} genarally use deep learning to assists some trees, or come up with a neural network structure so it resembles a tree.
A recent work \cite{wan2020nbdt} replaces the final fully connected layer of a neural network with a decision tree. Since the backbone features are still that of neural networks, the explanation is sought to be achieved via providing a means to humans to validate the decision as a good or bad one, rather than a complete logical reasoning of the decision.

In this paper, we show that any neural network having any activations has a directly equivalent decision tree representation. 
Thus, the induced tree output is exactly the same with that of the neural network and tree representation doesn't limit or require altering of the neural architecture in any way.
We believe that the decision tree equivalence provides better understanding of neural networks and paves the way to tackle the black-box nature of neural networks, e.g. via analyzing the category that a test sample belongs to, which can be extracted by the node rules that a sample is categorized.  
We show that the decision tree equivalent of a neural network can be found for either fully connected or convolutional neural networks which may include skip layers and normalizations as well.
Besides the interpretability aspect, we show that the induced tree is also advantageous to the corresponding neural network in terms of computational complexity, at the expense of increased storage memory.

Upon writing this paper, we have noticed the following works having overlaps with ours \cite{zhang2018tropical}, \cite{balestriero2018mad}, \cite{nguyen2020towards}, \cite{sudjianto2020unwrapping}, especially for feedforward ReLU networks. 
We extend the findings in these works to any activation function and also recurrent neural networks.

\section{Decision Tree Analysis of Neural Networks}
The derivations in this section will be first made for feedforward neural networks with piece-wise linear activation functions such as ReLU, Leaky ReLU, etc. 
Next, the analysis will be extended to any neural network with any activation function.

\label{sec:TREE}
\subsection{Fully Connected Networks}
\label{sec:FCN}
Let $\textbf{W}_i$ be the weight matrix of a network's $i^{th}$ layer.
Let $\sigma$ be any piece-wise linear activation function, and $\textbf{x}_0$ be the input to the neural network.
Then, the output and an intermediate feature of a feed-forward neural network can be represented as in Eq. \ref{eq: ffd}.

\begin{equation}
\label{eq: ffd}
\begin{split}
NN(\textbf{x}_{0})=\textbf{W}_{n-1}^T\sigma(\textbf{W}_{n-2}^T\sigma(...\textbf{W}_1^T\sigma(\textbf{W}_0^T\textbf{x}_{0})))
\\
x_{i}=\sigma(\textbf{W}_{i-1}^T\sigma(...\textbf{W}_1^T\sigma(\textbf{W}_0^T\textbf{x}_{0})))
\end{split}
\end{equation}

Note that in Eq. \ref{eq: ffd}, we omit any final activation (e.g. softmax) and we ignore the bias term as it can be simply included by concatenating a $1$ value to each $x_{i}$.
The activation function $\sigma$ acts as an element-wise scalar multiplication, hence the following can be written.

\begin{equation}
\label{eq: act}
\textbf{W}_{i}^T\sigma(\textbf{W}_{i-1}^T\textbf{x}_{i-1}) = \textbf{W}_{i}^T(\textbf{a}_{i-1}\odot(\textbf{W}_{i-1}^T\textbf{x}_{i-1}))
\end{equation}

In Eq. \ref{eq: act}, $\textbf{a}_{i-1}$ is a vector indicating the slopes of activations in the corresponding linear regions where $\textbf{W}_{i-1}^T\textbf{x}_{i-1}$ fall into, $\odot$ denotes element-wise multiplication.
Note that, $\textbf{a}_{i-1}$ can directly be interpreted as a categorization result since it includes indicators (slopes) of linear regions in activation function.
The Eq. \ref{eq: act} can be re-organized as follows.

\begin{equation}
\label{eq: act_matrix}
\textbf{W}_{i}^T\sigma(\textbf{W}_{i-1}^T\textbf{x}_{i-1}) = (\textbf{W}_{i}\odot\textbf{a}_{i-1})^T\textbf{W}_{i-1}^T\textbf{x}_{i-1}
\end{equation}

In Eq. \ref{eq: act_matrix}, we use $\odot$ as a column-wise element-wise multiplication on $\textbf{W}_i$.
This corresponds to element-wise multiplication by a matrix obtained via by repeating $\textbf{a}_{i-1}$ column-vector to match the size of $\textbf{W}_i$.
Using Eq. \ref{eq: act_matrix},  Eq. \ref{eq: ffd} can be rewritten as follows.

\begin{equation}
\label{eq: ffd_linear}
\begin{split}
NN(\textbf{x}_{0})=(\textbf{W}_{n-1}\odot\textbf{a}_{n-2})^T(\textbf{W}_{n-2}\odot\textbf{a}_{n-3})^T
\\
...(\textbf{W}_{1}\odot\textbf{a}_{0})^T\textbf{W}_{0}^T\textbf{x}_{0}
\\
\end{split}
\end{equation}

From Eq. \ref{eq: ffd_linear}, one can define an effective weight matrix $\hat{\textbf{W}}_{i}^T$ of a layer $i$ to be applied directly on input $\textbf{x}_0$ as follows:

\begin{equation}
\label{eq: eff_filter}
\begin{split}
{}_{\textbf{C}_{i-1}}\hat{\textbf{W}}_{i}^T=(\textbf{W}_{i}\odot\textbf{a}_{i-1})^T...(\textbf{W}_{1}\odot\textbf{a}_{0})^T\textbf{W}_{0}^T
\\
{}_{\textbf{C}_{i-1}}\hat{\textbf{W}}_{i}^T\textbf{x}_0=\textbf{W}_{i}^T\textbf{x}_i
\end{split}
\end{equation}

In Eq. \ref{eq: eff_filter}, the categorization vector until layer $i$ is defined as follows: $\textbf{c}_{i-1}=\textbf{a}_0\mathbin\Vert\textbf{a}_1\mathbin\Vert...\textbf{a}_{i-1}$, where $ \mathbin\Vert\ $ is the concatenation operator.  

One can directly observe from Eq. \ref{eq: eff_filter} that, the effective matrix of layer $i$ is only dependent on the categorization vectors from previous layers.
This indicates that in each layer, a new efficient filter is selected -to be applied on the network input- based on the previous categorizations/decisions.
This directly shows that a fully connected neural network can be represented as a single decision tree, where effective matrices acts as categorization rules. 
In each layer $i$, response of effective matrix ${}_{\textbf{C}_{i-1}}\hat{\textbf{W}}_{i}^T$ is categorized into $\textbf{a}_i$ vector, and based on this categorization result, next layer's effective matrix ${}_{\textbf{C}_{i}}\hat{\textbf{W}}_{i+1}^T$ is determined. 
A layer $i$ is thus represented as $k^{m_i}$-way categorization, where $m_i$ is the number filters in each layer $i$ and $k$ is the total number of linear regions in an activation.
This categorization in a layer $i$ can thus be represented by a tree of depth $m_i$, where a node in any depth is branched into $k$ categorizations.

In order to better illustrate the equivalent decision tree of a neural network, in Algorithm \ref{algo: eff_filter_eqn}, we rewrite Eq. \ref{eq: eff_filter} for the entire network, as an algorithm.
For the sake of simplicity and without loss of generality, we provide the algorithm with the ReLU activation function, where $a \in \{0,1\}$.
It can be clearly observed that, the lines $5-9$ in Algorithm \ref{algo: eff_filter_eqn} corresponds to a node in the decision tree, where a simple yes/no decision is made.

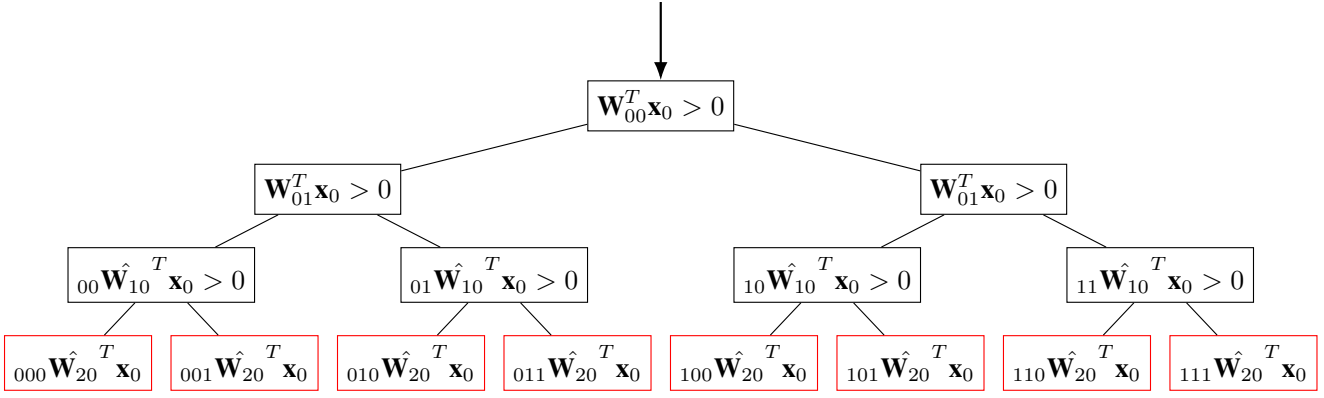
\begin{figure*}
\begin{adjustbox}{width=\linewidth}
\begin{forest} 
tikzQtree
[
$\textbf{W}_{00}^T\textbf{x}_0>0$, tikz={\draw[{Latex}-, thick] (.north) --++ (0,1);}
    [
    $\textbf{W}_{01}^T\textbf{x}_0>0$ 
			[
			${}_{00}\hat{\textbf{W}_{10}}^T\textbf{x}_0>0$   
				[
				${}_{000}\hat{\textbf{W}_{20}}^T\textbf{x}_0$   
				]
				[
				${}_{001}\hat{\textbf{W}_{20}}^T\textbf{x}_0$   
				]
			]
			[
			${}_{01}\hat{\textbf{W}_{10}}^T\textbf{x}_0>0$   
				[
				${}_{010}\hat{\textbf{W}_{20}}^T\textbf{x}_0$   
				]
				[
				${}_{011}\hat{\textbf{W}_{20}}^T\textbf{x}_0$   
				]
			]
    ]
    [
    $\textbf{W}_{01}^T\textbf{x}_0>0$ 
			[
			${}_{10}\hat{\textbf{W}_{10}}^T\textbf{x}_0>0$   
				[
				${}_{100}\hat{\textbf{W}_{20}}^T\textbf{x}_0$   
				]
				[
				${}_{101}\hat{\textbf{W}_{20}}^T\textbf{x}_0$   
				]
			]
			[
			${}_{11}\hat{\textbf{W}_{10}}^T\textbf{x}_0>0$   
				[
				${}_{110}\hat{\textbf{W}_{20}}^T\textbf{x}_0$   
				]
				[
				${}_{111}\hat{\textbf{W}_{20}}^T\textbf{x}_0$   
				]
			]
    ]
]
\end{forest}
\end{adjustbox}
\caption{Decision Tree for a 2-layer ReLU Neural Network}
\label{fig:2LayerTree}
\end{figure*}

The decision tree equivalent of a neural network can thus be constructed as in Algorithm \ref{algo: convertt}.
Using this algorithm, we share a a tree representation obtained for a neural network with three layers, having 2,1 and 1 filter for layer 1, 2 and 3 respectively. 
The network has ReLU activation in between layers, and no activation after last layer.
It can be observed from Algorithm \ref{algo: convertt} and Fig. \ref{fig:2LayerTree} that the depth of a NN-equivalent tree is $d=\sum_{i=0}^{n-2}{m_i}$, and total number of categories in last branch is $2^d$.
At first glance, the number of categories seem huge. 
For example, if first layer of a neural network contains 64 filters, there would exist at least $2^{64}$ branches in a tree, which is already intractable.
But, there may occur violating and redundant rules that would provide lossless pruning of the NN-equivalent tree.
Another observation is that, it is highly likely that not all categories will be realized during training due to the possibly much larger number of categories (tree leaves) than training data.
These categories can be pruned as well based on the application, and the data falling into these categories can be considered invalid, if the application permits.
In the next section, we show that such redundant, violating and unrealized categories indeed exist, by analysing decision trees of some neural networks.
But before that, we show that the tree equivalent of a neural network exists for skip connections, normalizations, convolutions, other activation functions and recurrence.

\subsubsection{Skip Connections}
We analyse a residual neural network of the following type:

\begin{equation}
\begin{gathered}
{}_{r}\textbf{x}_0=\textbf{W}_{0}^T\textbf{x}_0\\[1ex]
{}_{r}\textbf{x}_i={}_{r}\textbf{x}_{i-1}+\textbf{W}_{i}^T\sigma({}_{r}\textbf{x}_{i-1})
\end{gathered} \label{eq: residual}
\end{equation}
Using Eq. \ref{eq: residual}, via a similar analysis in Sec. \ref{sec:FCN}, one can rewrite ${}_{r}\textbf{x}_i$ as follows.

\begin{equation}
\begin{gathered}
{}_{r}\textbf{x}_i={}_{\textbf{a}_{i-1}}\hat{\textbf{W}}_{i}^T{}_{r}\textbf{x}_{i-1}   
\\[1ex]
{}_{\textbf{a}_{i-1}}\hat{\textbf{W}}_{i}^T=\textbf{I}+(\textbf{W}_{i}\odot\textbf{a}_{i-1})^T
\end{gathered} \label{eq: eff_residual_1}
\end{equation}

Finally, using ${}_{\textbf{a}_{i-1}}\hat{\textbf{W}}_{i}^T$ in Eq. \ref{eq: eff_residual_1}, one can define effective matrices for residual neural networks as follows.

\begin{equation}
\begin{gathered}
{}_{r}\textbf{x}_i={}_{r}\hat{\textbf{W}}_{i}^T\textbf{x}_0
\\[1ex]
{}_{r}\hat{\textbf{W}}_{i}^T= {}_{\textbf{a}_{i-1}}\hat{\textbf{W}}_{i}^T{}_{\textbf{a}_{i-2}}\hat{\textbf{W}}_{i-1}^T...{}_{\textbf{a}_{0}}\hat{\textbf{W}}_{1}^T\textbf{W}_{0}^T
\end{gathered} \label{eq: eff_residual_2}
\end{equation}
One can observe from  Eq. \ref{eq: eff_residual_2} that, for layer $i$, the residual effective matrix ${}_{r}\hat{\textbf{W}}_{i}^T$ is defined solely based on categorizations from the previous activations.
Similar to the analysis in Sec. \ref{sec:FCN}, this enables a tree equivalent of residual neural networks.

\begin{algorithm}[t]
$\hat{\textbf{W}}=\textbf{W}_{0}$ 
\\
\For{$i=0$ \KwTo $n-2$}
{   
$\textbf{a}=[]$
\\
\For{$j=0$ \KwTo $m_i-1$}
{   
  \uIf{$\hat{\textbf{W}}_{ij}^T\textbf{x}_0>0$}{
    $\textbf{a}.append(1)$
  }
  \Else{
    $\textbf{a}.append(0)$
  }
}
$\hat{\textbf{W}}=\hat{\textbf{W}}(\textbf{W}_{i+1}\odot\textbf{a})$
}
\Return $\hat{\textbf{W}}^T\textbf{x}_0$
\caption{Algorithm of Eq. \ref{eq: eff_filter} for ReLU networks}
\label{algo: eff_filter_eqn}
\end{algorithm}

\begin{algorithm}[t]
Initialize Tree: Set root.
\\
Branch all leafs to $k$ nodes, decision rule is first effective filter.
\\
Branch all nodes to $k$ more nodes, and repeat until all effective filters in a layer is covered.
\\
Calculate effective matrix for each leaf via Eq. \ref{eq: eff_filter}. Repeat 2,3.
\\
Repeat 4 until all layers are covered.
\\
\Return Tree
\caption{Algorithm of converting neural networks to decision trees}
\label{algo: convertt}
\end{algorithm}

\subsubsection{Normalization Layers}

A separate analysis is not needed for any normalization layer, as popular normalization layers are linear, and after training, they can be embedded into the linear layer that it comes after or before, in pre-activation or post-activation normalizations respectively. 

\begin{figure*}
\begin{adjustbox}{width=\linewidth}
\begin{forest} 
tikzQtree
[
$x<-1.16$, tikz={\draw[{Latex}-, thick] (.north) --++ (0,1);}
    [
    $x<0.32$ 
			[
			$x>1$   
				[
				$x<-0.1$
				[${}_{0}\hat{\textbf{W}}^T\textbf{x}_0$]
				[${}_{1}\hat{\textbf{W}}^T\textbf{x}_0$]
				]
				[
				$\textbf{x}_0<-0.1$
				[${}_{2}\hat{\textbf{W}}^Tx$]
				[${}_{3}\hat{\textbf{W}}^Tx$]
				]
			]
			[
			$x>0.54$
				[
				$x<0.11$
				[${}_{4}\hat{\textbf{W}}^Tx$]
				[${}_{5}\hat{\textbf{W}}^Tx$]
				] 
				[
				$x<0.11$
				[${}_{6}\hat{\textbf{W}}^Tx$]
				[${}_{7}\hat{\textbf{W}}^Tx$]
				] 
			]
    ]
    [
    $x<0.32$
			[
			$x>0.52$   
				[
				$x<-0.7$
				[${}_{8}\hat{\textbf{W}}^Tx$]
				[${}_{9}\hat{\textbf{W}}^Tx$]
				]
				[
				$x<-0.7$
				[${}_{10}\hat{\textbf{W}}^Tx$]
				[${}_{11}\hat{\textbf{W}}^Tx$]
				]
			]
			[
			$x>0.39$
				[
				$x<-0.38$
				[${}_{12}\hat{\textbf{W}}^Tx$]
				[${}_{13}\hat{\textbf{W}}^Tx$]
				] 
				[
				$x<-0.38$
				[${}_{14}\hat{\textbf{W}}^Tx$]
				[${}_{15}\hat{\textbf{W}}^Tx$]
				] 
			]
    ]
]
\end{forest}
\end{adjustbox}
\caption{Decision Tree for a $y=x^2$ Regression Neural Network}
\label{fig:RegressTree}
\end{figure*}
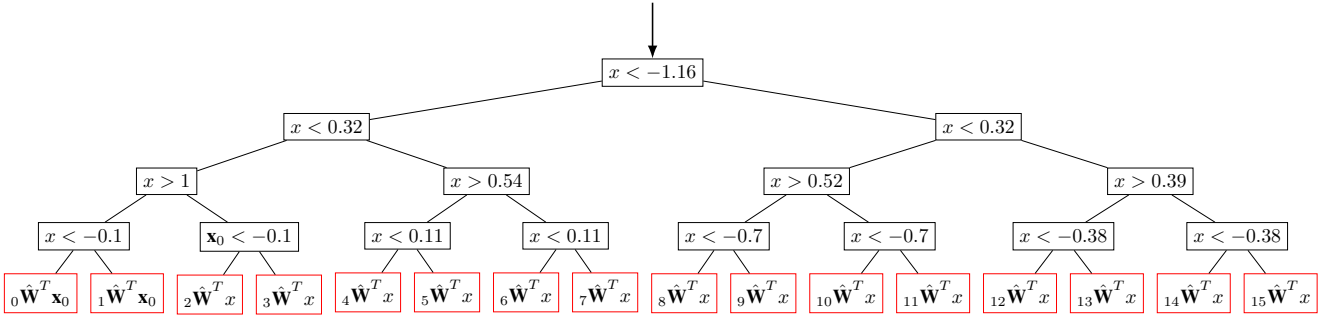

\subsection{Convolutional Neural Networks}

Let $\textbf{K}_i : C_{i+1}\times C_i\times M_i\times N_i $ be the convolution kernel for layer i, applying on an input $\textbf{F}_{i} : C_i\times H_i\times W_i$.
Note that $M_i$ and $N_i$ denote the spatial size of the convolutional kernel, and $H_i$ and $W_i$ denote the spatial size of the input.

One can write the output of a convolutional neural network $CNN(\textbf{F}_0)$, and an intermediate feature $\textbf{F}_{i}$ as follows.

\begin{equation}
\label{eq: convnet}
\begin{split}
CNN(\textbf{F}_0)=\textbf{K}_{n-1}*\sigma(\textbf{K}_{n-2}*\sigma(...\sigma(\textbf{K}_{0}*\textbf{F}_0))
\\
\textbf{F}_{i}=\sigma(\textbf{K}_{i-1}*\sigma(...\sigma(\textbf{K}_{0}*\textbf{F}_0))
\end{split}
\end{equation}

Similar to the fully connected network analysis, one can write the following, due to element-wise scalar multiplication nature of the activation function.

\begin{equation}
\label{eq: convnetact}
\textbf{K}_{i}*\sigma(\textbf{K}_{i-1}*\textbf{F}_{i-1}) = (\textbf{K}_{i}\odot\textbf{a}_{i-1})*(\textbf{K}_{i-1}*\textbf{F}_{i-1})
\end{equation}

In Eq. \ref{eq: convnetact}, $\textbf{a}_{i-1}$ is of same spatial size as $\textbf{K}_{i}$ and consists of the slopes of activation function in corresponding regions in the previous feature $\textbf{F}_{i-1}$.
Note that the above only holds for a specific spatial region, and there exists a separate $\textbf{a}_{i-1}$ for each spatial region that the convolution $\textbf{K}_{i-1}$ is applied to.
For example, if $\textbf{K}_{i-1}$ is a $3\times 3$ kernel, there exists a separate $\textbf{a}_{i-1}$ for all $3\times 3$ regions that the convolution is applied to. 
An effective convolution ${}_{\textbf{C}_{i-1}}\hat{\textbf{K}}_{i}$ can be written as follows.

\begin{equation}
\label{eq: eff_conv}
\begin{split}
{}_{\textbf{c}_{i-1}}\hat{\textbf{K}}_{i}=(\textbf{K}_{i}\odot\textbf{a}_{i-1})*...*(\textbf{K}_{1}\odot\textbf{a}_{0})*\textbf{K}_{0}
\\
{}_{\textbf{c}_{i-1}}\hat{\textbf{K}}_{i}*\textbf{x}_0=\textbf{K}_{i}*\textbf{x}_i
\end{split}
\end{equation}

Note that in Eq. \ref{eq: eff_conv}, ${}_{\textbf{C}_{i-1}}\hat{\textbf{K}}_{i}$ contains specific effective convolutions per region, where a region is defined according to the receptive field of layer $i$. 
$\textbf{c}$ is defined as the concatenated categorization results of all relevant regions from previous layers.

One can observe from Eq. \ref{eq: eff_conv} that effecive convolutions are only dependent on categorizations coming from activations, which 
enables the tree equivalence -similar to the analysis for fully connected network.
A difference from fully connected layer case is that many decisions are made on partial input regions rather than entire $\textbf{x}_0$.

\begin{figure*}
\centering
\subfloat[\centering Cleaned Tree\label{fig:CleanTree}]{
\begin{forest} 
tikzQtree
[
$x<-1.16$, tikz={\draw[{Latex}-, thick] (.north) --++ (0,1);}
	[
	$x<0.32$
		[
		$x>1$
			[$0.55 \textbf{x}+0.09$]
			[$3.47 \textbf{x}-2.83$]
		]
		[
		$x<0.11.$
			[$2.37 \textbf{x}-0.50$]
			[$-1.00 \textbf{x}-0.12$]
		]
	] 
	[
	$-3.67\textbf{x}-3.22$
	]
] 
\end{forest}
}
\qquad
\subfloat[\centering Neural Network Approximation of $y=x^2$ \label{fig:regressplot}]{{\includegraphics[width=7cm]{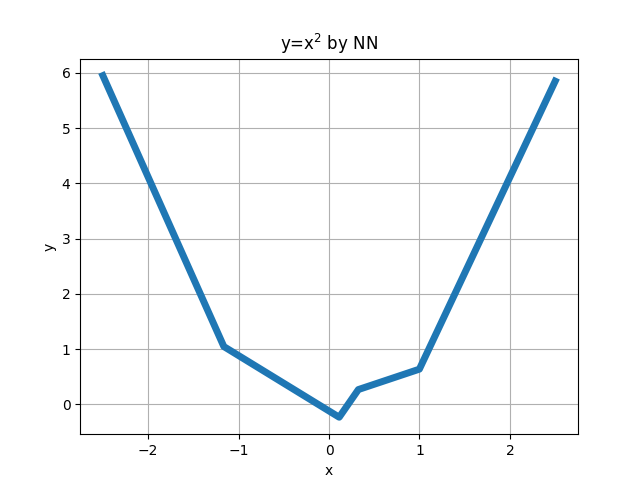} }}%
\caption{Cleaned Decision Tree for a $y=x^2$ Regression Neural Network}
\label{fig:RegressTreeClean}
\end{figure*}
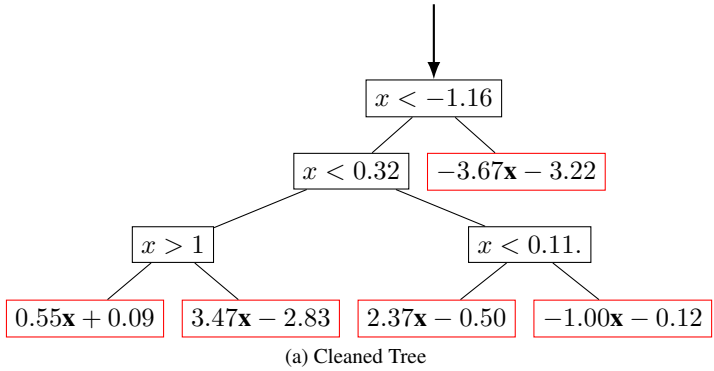
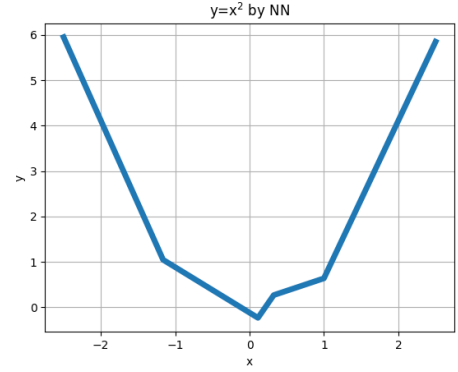

\begin{figure*}
\begin{adjustbox}{width=\linewidth}
\begin{forest} 
tikzQtree
[
$-0.98x-0.49y+0.95$, tikz={\draw[{Latex}-, thick] (.north) --++ (0,1);}
    [
    $-1.6x+0.2y-0.07$ 
			[
			$-0.15x+0.19y+0.41$   
				[
				$1$
				]
				[
				$0.45x-0.3y+0.05$
					[$0$]
					[$1.6x-1.35y-1.44$
						[0]
						[1]					
					]
				]
			]
			[
			$0$
			]
    ]
    [
    $-1.6x+0.2y-0.07$
			[
			$0.5x+0.52y-0.22$   
				[
				$1$
				]
				[
				$-0.46x-0.76y+0.94$
				[$0$]
				[$-2.72x-3.53y+2.77$
				[0]
				[1]				
				]
				]
			]
			[
			$-0.5x+0.64y-0.27$
				[
				$1.51x-1y+1.03$
				[$1.59x-1.35y+0.78$
				[0]
				[1]
				]
				[$4.18x-3.17y+2.54$
				[0]
				[1]				
				]
				] 
				[
				$1.51x-1y+1.03$
				[$0$]
				[$5.31x-4.51y+3.14$
				[0]
				[1]				
				]
				] 
			]
    ]
]
\end{forest}
\end{adjustbox}
\caption{Classification Tree for a Half-Moon Classification Neural Network}
\label{fig:ClassTree}
\end{figure*}
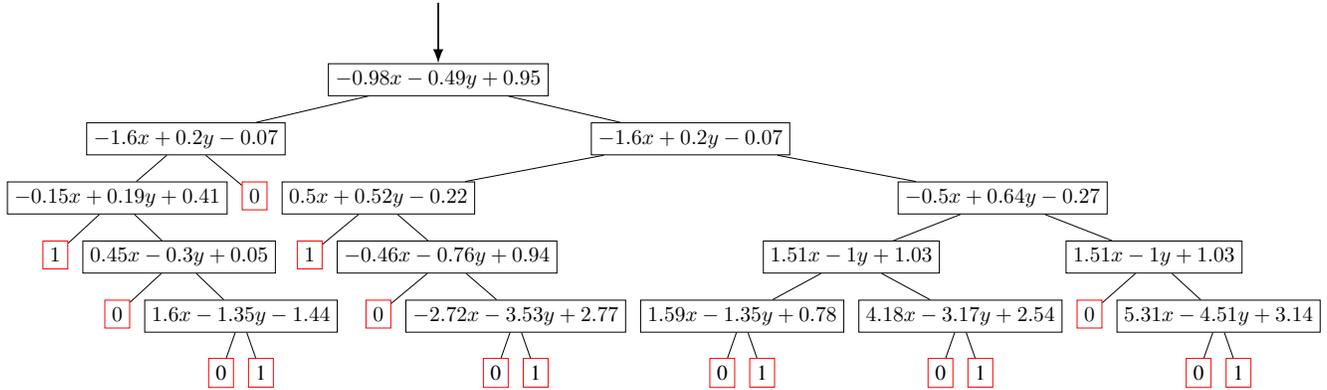

\begin{figure}%
    \centering
    \label{fig:halfmoon}
    {{\includegraphics[width=7cm]{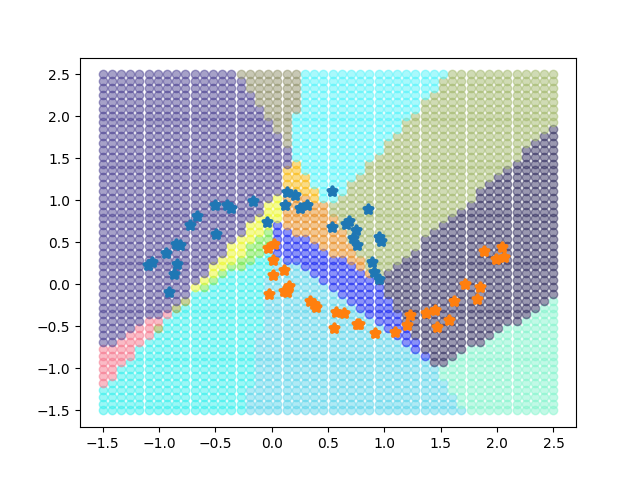} }}%
    \caption{Categorizations made by the decision tree for half-moon dataset}
\label{fig:HalfMoon}%
\end{figure}

\subsection{Continuous Activation Functions}
\label{contact}
In Eq. \ref{eq: act}, for piece-wise linear activations, elements of \textbf{a} can have a number of values limited by the piece-wise linear regions in the activation function.
This number defines the number of child nodes per effective filter.
The extension to continuous activation functions is trivial as they can be considered as piece-wise linear functions with infinite regions.
Therefore, for continuous activations, the neural network equivalent tree immediately becomes infinite width even for a single filter.
This might not be a useful result, but we provide this discussion here for completeness.
In order to guarantee finite trees, one may consider using quantized versions of continuous activations which may result in a few piece-wise linear regions, hence few child nodes per activation.

\subsection{Recurrent Networks}

As recurrent neural networks (RNNs) can be unrolled to feed-forward representation, RNNs can also be equivalently represented as decision trees.
We study following recurrent neural network.
Note that we simply omit the bias terms as they can be represented by concatenating a 1 value to input vectors.

\begin{equation}
\label{eq: rnn}
\begin{split}
\textbf{h}^{(t)}=\sigma(\textbf{W}^T\textbf{h}^{(t-1)}+\textbf{U}^T\textbf{x}^{(t)})
\\
\textbf{o}^{(t)}=\textbf{V}^T\textbf{h}^{(t)}
\end{split}
\end{equation} 

Similar to previous analysis, one can rewrite $\textbf{h}^{(t)}$ as follows.

\begin{equation}
\label{eq: rnn_linact}
\textbf{h}^{(t)}=\textbf{a}^{(t)}\odot(\textbf{W}^T\textbf{h}^{(t-1)}+\textbf{U}^T\textbf{x}^{(t)})
\end{equation} 

Eq. \ref{eq: rnn_linact} can be rewritten follows.
\begin{equation}
\label{eq: rnn_linact_2}
\begin{split}
\textbf{h}^{(t)}=\textbf{a}^{(t)}\odot(\prod_{j=(t-1)}^{1}(\textbf{W}^T\odot\textbf{a}^{(j)}))\textbf{W}^T\textbf{h}^{(0)}
\\
+\textbf{a}^{(t)}\odot\sum_{i=1}^{t}(\prod_{j=(t-1)}^{i}(\textbf{W}^T\odot\textbf{a}^{(j)}))\textbf{U}^T\textbf{x}^{(i)}
\end{split}
\end{equation} 

Note that in Eq. \ref{eq: rnn_linact_2}, the product operator stands for matrix multiplication, its steps are $-1$ and we consider the output of product operator to be 1 when $i=t$.
One can rewrite Eq. \ref{eq: rnn_linact_2} by introducing ${}_{\textbf{c}_{j}}\hat{\textbf{W}}_{j}$ as follows.

\begin{equation}
\label{eq: rnn_linact_3}
\begin{split}
\textbf{h}^{(t)}=\textbf{a}^{(t)}\odot{}_{\textbf{c}_{1}}\hat{\textbf{W}}_{1}\textbf{W}^T\textbf{h}^{(0)}+\textbf{a}^{(t)}\odot\sum_{i=1}^{t}{}_{\textbf{c}_{i}}\hat{\textbf{W}}_{i}\textbf{U}^T\textbf{x}^{(i)}
\\
{}_{\textbf{c}_{i}}\hat{\textbf{W}}_{i}^T=\prod_{j=(t-1)}^{i}(\textbf{W}^T\odot\textbf{a}^{(j)})
\end{split}
\end{equation} 

Combining Eq. \ref{eq: rnn_linact_3} and Eq. \ref{eq: rnn}, one can write $\textbf{o}^{(t)}$ as follows.

\begin{equation}
\label{eq: rnn_output}
\textbf{o}^{(t)}={}_{\textbf{a}^{(t)}}\hat{\textbf{V}}^T{}_{\textbf{c}_{1}}\hat{\textbf{W}}_{1}\textbf{W}^T\textbf{h}^{(0)}+{}_{\textbf{a}^{(t)}}\hat{\textbf{V}}^T\sum_{i=1}^{t}{}_{\textbf{c}_{i}}\hat{\textbf{W}}_{i}\textbf{U}^T\textbf{x}^{(i)}
\end{equation} 

Eq. \ref{eq: rnn_output} can be further simplified to the following.

\begin{equation}
\label{eq: rnn_output2}
\textbf{o}^{(t)}={}_{\textbf{c}_{1}}\hat{\textbf{Z}}_{1}^T\textbf{W}^T\textbf{h}^{(0)}+\sum_{i=1}^{t}{}_{\textbf{c}_{i}}\hat{\textbf{Z}}_{i}\textbf{U}^T\textbf{x}^{(i)}
\end{equation} 

In Eq. \ref{eq: rnn_output2}, ${}_{\textbf{c}_{i}}\hat{\textbf{Z}}_{i}^T ={}_{\textbf{a}^{(t)}}\hat{\textbf{V}}^T{}_{\textbf{c}_{i}}\hat{\textbf{W}}_{i}$ .As one can observe from Eq. \ref{eq: rnn_output2}, the RNN output only depends on the categorization vector $\textbf{c}_{i}$, which enables the tree equivalence -similar to previous analysis.

Note that for RNNs, a popular choice for $\sigma$ in Eq. \ref{eq: rnn} is $tanh$. As mentioned in Section \ref{contact}, in order to provide finite trees, one might consider using a piece-wise linear approximation of $tanh$.

\section{Experimental Results}
First, we make a toy experiment where we fit a neural network to: $y=x^2$ equation. 
The neural network has $3$ dense layers with $2$ filters each, except for last layer which has 1 filter.
The network uses leaky-ReLU activations after fully connected layers, except for the last layer which has no post-activation.
We have used negative slope of $0.3$ for leaky-ReLU which is the default value in Tensorflow \cite{tensorflow2015-whitepaper}.
The network was trained with $5000$ $(x,y)$ pairs where $x$ was regularly sampled from $[-2.5,2.5]$ interval.
Fig. \ref{fig:RegressTree} shows the decision tree corresponding to the neural network.
In the tree, every black rectangle box indicates a rule, left child from the box means the rule does not hold, and the right child means the rule holds.
For better visualization, the rules are obtained via converting $\textbf{w}^Tx+\beta>0$ to direct inequalities acting on $x$. 
This can be done for the particular regression $y=x^2$, since $x$ is a scalar.
In every leaf, the network applies a linear function -indicated by a red rectangle- based on the decisions so far.
We have avoided writing these functions explicitly due to limited space.
At first glance, the tree representation of a neural network in this example seems large due to the $2^{\sum_i^{n-2}m_i}=2^4=16$ categorizations.
However, we notice that a lot of the rules in the decision tree is redundant, and hence some paths in the decision tree becomes invalid.
An example to redundant rule is checking $x<0.32$ after $x<-1.16$ rule holds.
This directly creates the invalid left child for this node.
Hence, the tree can be cleaned via removing the left child in this case, and merging the categorization rule to the stricter one : $x<-1.16$ in the particular case.
Via cleaning the decision tree in Fig. \ref{fig:RegressTree}, we obtain the simpler tree in Fig. \ref{fig:CleanTree}, which only consists of $5$ categories instead of $16$.
The $5$ categories are directly visible also from the model response in Fig. \ref{fig:regressplot}. 
The interpretation of the neural network is thus straightforward: for each region whose boundaries are determined via the decision tree representation, the network approximates the non-linear $y=x^2$ equation by a linear equation.
One can clearly interpret and moreover make deduction from the decision tree, some of which are as follows.
The neural network is unable to grasp the symmetrical nature of the regression problem which is evident from the fact that the decision boundaries are asymmetrical. 
The region in below $-1.16$ and above $1$ is unbounded and thus neural decisions lose accuracy as x goes beyond these boundaries.

Next, we investigate another toy problem of classifying half-moons and analyse the decision tree produced by a neural network.
We train a fully connected neural network with 3 layers with leaky-ReLU activations, except for last layer which has sigmoid activation.
Each layer has $2$ filters except for the last layer which has $1$. 
The cleaned decision tree induced by the trained network is shown in Fig. \ref{fig:ClassTree}.
The decision tree finds many categories whose boundaries are determined by the rules in the tree, where each category is assigned a single class.
In order to better visualize the categories, we illustrate them with different colors in Fig. \ref{fig:HalfMoon}.
One can make several deductions from the decision tree such as some regions are very well-defined, bounded and the classifications they make are perfectly in line with the training data, thus  these regions are very reliable.
There are unbounded categories which help obtaining accurate classification boundaries, yet fail to provide a compact representation of the training data, these may correspond to inaccurate extrapolations made by neural decisions.
There are also some categories that emerged although none of the training data falls to them.

\begin{table}
\begin{adjustbox}{width=\linewidth}
\begin{tabular}{|c|c|c|c|c|c|c|}
 \hline
   & \multicolumn{3}{|c|}{$y=x^2$} & \multicolumn{3}{|c|}{Half-Moon} \\
 \hline
   & Param. & Comp. & Mult./Add. & Param. & Comp. & Mult./Add. \\
 \hline
  Tree & 14 & 2.6 & 2 & 39 & 4.1 & 8.2 \\
 \hline
  NN &  13 & 4 & 16 & 15 & 5 & 25 \\
 \hline
\end{tabular}
\end{adjustbox}
\caption{\label{tab:comp}Computation and memory analysis of toy problems.}
\end{table}

Besides the interpretability aspect, the decision tree representation also provides some computational advantages.
In Table \ref{tab:comp}, we compare the number of parameters, float-point comparisons and multiplication or addition operations of the neural network and the tree induced by it.
Note that the comparisons, multiplications and additions in the tree representation are given as expected values, since per each category depth of the tree is different.
As the induced tree is an unfolding of the neural network, it covers all possible routes and keeps all possible effective filters in memory.
Thus, as expected, the number of parameters in the tree representation of a neural network is larger than that of the network.
In the induced tree, in every layer $i$, a maximum of $m_i$ filters are applied directly on the input, whereas in the neural network always $m_i$ filters are applied on the previous feature, which is usually much larger than the input in the feature dimension.
Thus, computation-wise, the tree representation is advantageous compared to the neural network one.

\section{Conclusion}

In this manuscript, we have shown that neural networks can be equivalently represented as decision trees.
The tree equivalence holds for fully connected layers, convolutional layers, residual connections, normalizations, recurrent layers and any activation.
We believe that this tree equivalence provides directions to tackle the black-box nature of neural networks.

{\small
\bibliographystyle{ieee_fullname}
\bibliography{egbib}
}

\end{document}